\documentclass{article}

\usepackage{microtype}
\usepackage{graphicx}
\usepackage{subfigure}
\usepackage{booktabs} 

\usepackage{multirow}
\usepackage{mathrsfs}
\usepackage{amsmath}

\usepackage{amssymb}

\usepackage{subfigure} 

\usepackage{amsthm}
\theoremstyle{definition}
\newtheorem{definition}{Definition}
\newtheorem{theorem}{Theorem}
\newtheorem{lemma}{Lemma}

\usepackage{hyperref}

\usepackage[accepted]{icml2021}

\begin{document}

\twocolumn[
\icmltitle{Learning Class Unique Features in Fine-Grained Visual Classification}



\icmlsetsymbol{equal}{*}

\begin{icmlauthorlist}
\icmlauthor{Runkai Zheng}{e,j}
\icmlauthor{Zhijia Yu}{t}
\icmlauthor{Yinqi Zhang}{j}
\icmlauthor{Chris Ding}{d}
\icmlauthor{Hei Victor Cheng}{u}
\icmlauthor{Li Liu}{c}
\end{icmlauthorlist}

\icmlaffiliation{e}{Elecholic, Guangzhou, China}
\icmlaffiliation{j}{Jinan University, Guangzhou, China}
\icmlaffiliation{t}{Tsinghua University, Beijing, China}
\icmlaffiliation{d}{The Chinese University of Hong Kong Shenzhen, Shenzhen, China}
\icmlaffiliation{c}{Shenzhen research institute of big data, the Chinese University of Hong Kong Shenzhen, Shenzhen, China}
\icmlaffiliation{u}{Department of Electrical and Computer Engineering, University of Toronto, Toronto, Canada}

\icmlcorrespondingauthor{Li Liu}{liuli@cuhk.edu.cn}

\icmlkeywords{Machine Learning, ICML}

\vskip 0.3in
]

\printAffiliations{}
\begin{abstract}

A major challenge in Fine-Grained Visual Classification (FGVC) is distinguishing various categories with high inter-class similarity by learning the feature that differentiate the details. Conventional cross entropy trained Convolutional Neural Network (CNN) fails this challenge as it may suffer from producing inter-class invariant features in FGVC. In this work, we innovatively propose to regularize the training of CNN by enforcing the uniqueness of the features to each category from an information theoretic perspective. To achieve this goal, we formulate a minimax loss based on a game theoretic framework, where a Nash equilibria is proved to be consistent with this regularization objective. Besides, to prevent from a feasible solution of minimax loss that may produce redundant features, we present a Feature Redundancy Loss (FRL) based on normalized inner product between each selected feature map pair to complement the proposed minimax loss. Superior experimental results on several influential benchmarks along with visualization show that our method gives full play to the performance of the baseline model without additional computation and achieves comparable results with state-of-the-art models.

\end{abstract}

\begin{figure}[tb]
    \centering
    \includegraphics[width=0.9\linewidth]{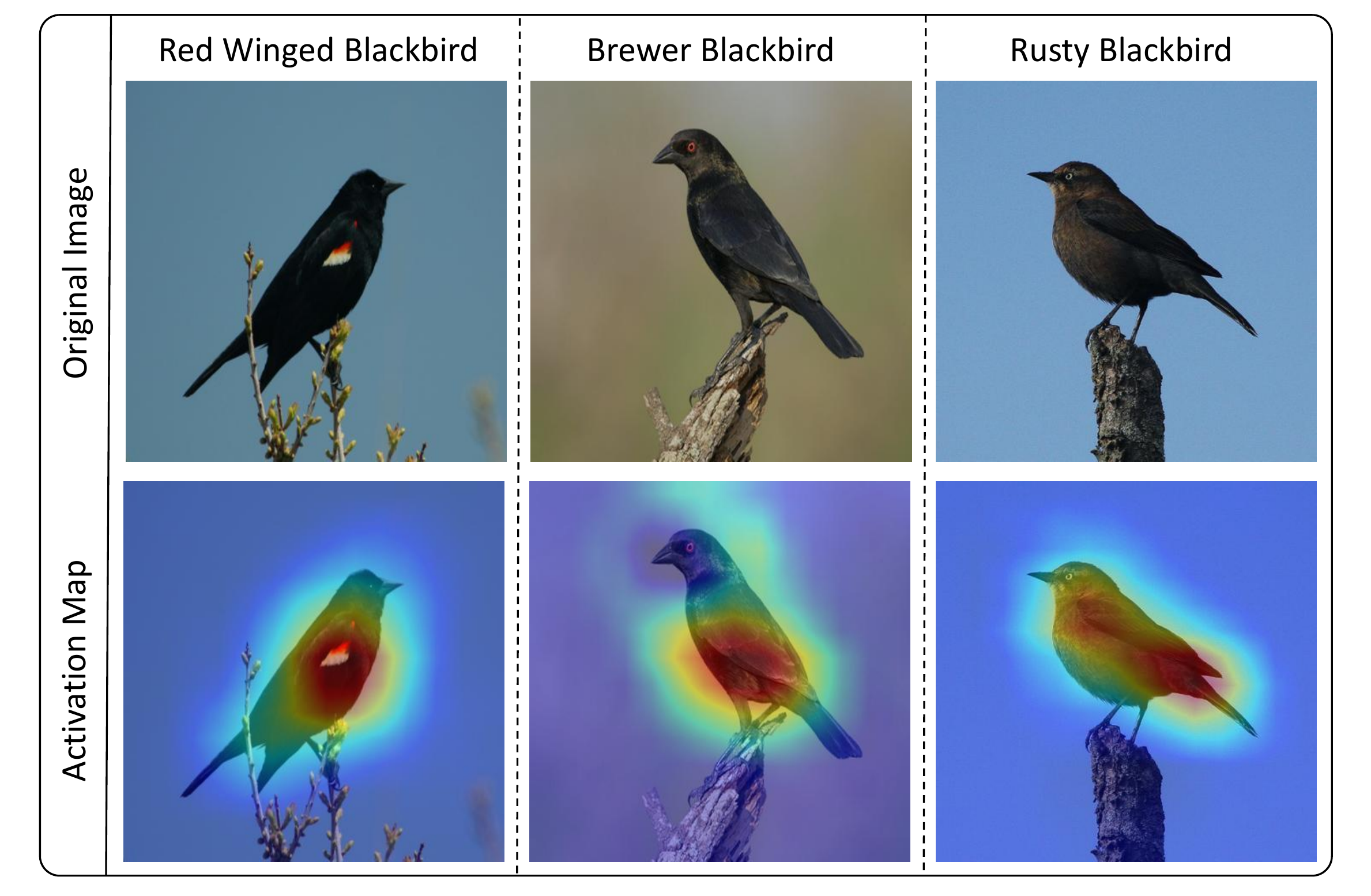}
    \caption{Three images of similar bird species from CUB-200-2011 dataset are selected to present the major challenge in FGVC. The first row presents the original images and the second row presents the activation maps. The activation maps are obtained from forward propagating the image of Red Winged Blackbird to the CE trained CNN and extracting the penultimate layer feature maps. Then the channel that has the largest mean activation value is selected. An observation is that the same channel also has high activation value when we input the other two categories of images (Brewer Blackbird and Rusty Blackbird).}
    \label{fig:intro}
\end{figure}

Convolutional Neural Networks (CNN) achieves a great success in the computer vision domain. The large diversity in standard visual recognition tasks make it possible for CNN to well learn discriminative features by minimizing the cross entropy (CE) loss. When it comes to Fine-Grained Visual Classification (FGVC), different categories with highly similar appearance leads to inter-class invariants and intra-class variants, which limit the performance of standard CE trained CNN. 

Over the past few years, FGVC has attracted lots of attention in research community. Early works \cite{zhang2014part,bransonbird,wei2016mask} utilize multi-stage architecture that consists of a localization network and a classification network. The localization network is responsible for detecting discriminative regions, which requires bounding box or part annotations for training. Then the classification network works on the cropped regions given by the localization network. However, these approaches depend on annotations and cannot be trained end-to-end, thus lead to extra training cost. To solve the above mentioned challenges, recent approaches manage to develop end-to-end networks that only require weak supervision. Commonly these approaches outperform baseline models by mimicking human actions like attention mechanism and part localization \cite{zheng2017learning,fu2017look,sun2018multi,wang2018learning,chen2019destruction,ding2019selective,wang2020graph}. However, these works focus on discriminative parts on spatial domain. Even if the regions of interest are correctly cropped or detected, CNN will inevitably encode unnecessary information that may mix up with other categories. As shown in Fig. \ref{fig:intro}, the problem is, although the model can well localize the discriminative parts, the channel that responsible for detecting the feature will also be activated when encounter a similar texture. That means localization is not sufficient for learning a good feature if the filters can not precisely encode the unique feature. It is a challenge to extract the features that contain unique information about the categories.

In this work, we propose an explicit regularization objective for encoding unique features with a theoretical guarantee. Our motivation is based on an \textbf{assumption} that a unique feature should contain only the information of a specific category and not any other categories. In other words, for a given image, we expect the extracted features to be highly correlated with the target class without extra information about the non-target classes. We call this kind of features Class Unique Features (CUFs). To achieve this goal, we formulate CUF using the Mutual Information (MI), from which we deduce an explicit regularization objective, i.e., \textit{Maximum Non-Target distribution Entropy (MaxNTE)}. To efficiently optimize the objective, we propose a game theoretic framework to simplify the problem formulation. Under this game theoretic framework, the existence of Nash equilibria and the consistency between the outcome and our objective are proved rigorously. 

In summary, our contribution includes:
\begin{enumerate}
\item We innovatively formulate our assumption to an ideal CUF learner from an information theoretic perspective and deduce an explicit regularization objective.

\item We construct a game-theoretic framework between the model and the adversary. On this basis, we arrive at a simple yet efficient minimax (MM) loss to achieve the regularization goal. To reduce the feature redundancy brought by the minimax loss, we further propose a Feature Redundancy Loss (FRL), encouraging the model to focus on multiple discriminative parts, as a complement to the minimax loss.

\item Experimental results on influential benchmarks of both FGVC and standard visual classification show that our method outperforms the baseline models by a large margin and achieves state-of-the-art (SOTA) results on FGVC-Aircraft and Standard Cars dataset.
\end{enumerate}

\section{Related Work}

\subsection{Fine-Grained Visual Classification}
Recently, FGVC is a research hotspot in the field of computer vision. We mainly discuss related works according to the following three research branches.

\textbf{Attention mechanism and part localization} were explored to settle this problem, as the model is able to learn to pay attention to the region or features that contain inter-class variations. Benefited from the interaction of part learning and feature learning, Multi-Attention CNN (MA-CNN) was proposed in \cite{zheng2017learning} to extract part-based fine-grained features. 
\cite{fu2017look} utilized recurrent architecture to repeatedly crop and scale the regions of interest by attention mechanism.
\cite{sun2018multi} proposed One-Squeeze Multi-Excitation that generate multiple attention map based on Multi-Attention Multi Class Constraint to efficiently obtain the highly discriminative part.
\cite{wang2018learning} used $1\times1$ convolution kernel as a discriminative patch detector and designed an asymmetric, multi-channel structure to enhance the learning of discriminative mid-level patches.
\cite{chen2019destruction} shuffled the local regions to enforce the network to focus on the most discriminative patches.
\cite{ding2019selective} used sparse attention for feature sampling to capture detailed visual evidence without losing the context information.

\textbf{High-order statistics} were explored for aggregating features to improve the first-order statistics such as max pooling and average pooling because they were difficult to capture the diversity of features among different categories.
\cite{lin2015bilinear} produced an image descriptor via pooling the outer product from two CNN feature extractor sub-branches, which was able to model local pairwise feature interactions in a translational invariant manner.
\cite{gao2016compact} approximated bilinear pooling operation by applying low-dimensional approximation of the polynomial kernel to speed up the computation.
\cite{wang2019deep} inserted Matrix Power Normalized COVariance (MPN-COV) block into the final layer of convolutions to obtain a global representation by second order statics. 

\textbf{Regularization based methods} usually do not need extra computation and thus are much light weight compared with the above mentioned methods. They developed efficient training manner that can boost the performance of simple baseline models. \cite{dubey2018maximum} formulated the relation between model selection and feature diversity, and utilizing the idea of maximum entropy to minimize the lower bound of Frobenius norm of the weights and thus improved the performance of models in fine-grained visual tasks. \cite{dubey2018pairwise} minimized the L2 distance between the prediction probability distribution of the random sample pairs of the training set to confuse the network and prevent from overfitting. Our method is also based on regularization of output distribution, and thus can be a simple and lightweight tool to be used among similar tasks.

\subsection{Label smoothing}
Label smoothing \cite{szegedy2016rethinking} was first proposed to prevent deep learning model from overconfident in classification problem. The characteristic of softmax function makes it impossible for a model to convergent to the hard 0 and 1 targets \cite{Goodfellow-et-al-2016}. Thus the model may keep seeking for extreme prediction and become overfitting. Label smoothing introduces uniform noise distribution $u$ to the ground truth labels by replacing 0 and 1 targets with $\frac{\epsilon}{n-1}$ and $1-\epsilon$, where $\epsilon$ is a hyperparameter determines the amount of smoothing. Our method produces uniformly distributed probabilities on non-target class, which is similar to the ground truth of label smoothing. However, our method is different from label smoothing in terms of motivation, training manner and resulting outputs. 
More precisely, 1) our motivation is to achieve an assumption on MI between extracted features and output distribution, while LS is proposed to inject noise to the labels to prevent from extreme logits and overfitting.
2) We use a minimax loss to achieve our objective while LS directly takes the designed target to supervise the model.
3) In terms of the regularization results, our method leads to uniform distribution on non-target classes, but LS can not.

\subsection{Mutual Information}
Mutual Information (MI) is a measure of information in information theory, which indicates the mutual dependencies between two random variables. More specifically, it quantifies the amount of information about another random variable when one of the variables is observed. 
MI has been used in the field of deep learning. 
\cite{tishby2015deep} firstly showed that Deep Neural Networks (DNN) can be quantified by the mutual information between the layers and the input and output variables. They provide a novel perspective that the goal of DNN is to optimize Information Bottleneck (IB) trade off between compression and prediction. 
\cite{hjelm2018learning} wielded the rich knowledge about mutual information into the construction of encoder, called Deep InforMax (DIM), which maximized the mutual information between the inputs and the high-level representation. Belghazi et al. \cite{belghazi2018mutual} proposed Mutual Information Neural Estimator (MINE) and applied it to Generative Adversarial Networks (GANs) 
\cite{goodfellow2014generative} to improve the reconstruction quality and alleviate mode-drop in GANs. 
\subsection{Game Theory}
Game Theory provides mathematical models of strategic interaction among intelligent decision makers \cite{myerson1991game}. It is a mathematical theory and method to study the phenomenon of competition, and was studied in \cite{osborne2004introduction} the interaction between the formulaic incentive structures. With the increasing popularity of Artificial Intelligent, game theory has been applied to different fields including Multi-Agent Reinforcement Learning \cite{bowling2000analysis} and GANs \cite{goodfellow2014generative, goodfellow2016nips}. In this work, we apply the game theory to the object recognition task.

\section{Method}
Our goal is to extract features that do not contain information from non-target classes, i.e., CUF. This can be formulated by minimizing the MI between extracted feature and the non-target output distribution. By the formulation, we further deduce a regularization objective, where a key finding is derived (i.e., all output probabilities over non-target classes should be uniformly distributed). To efficiently optimize this objective function, we propose a minimax loss to simplify the reformulated regularization objective. 

\subsection{Mutual Information based Problem Formulation}

 We denote the input space by $\mathcal{X}$, the label space by $C=\{1, 2, \dots, n \}$, where $n>2$ is the number of classes. 
 The training data are all i.i.d sample pairs $(x, y)$, where $y$ is in the form of a $n$-dimensional one-hot vector.

Let $\Phi$ be a parametric function mapping from the input space to the feature space. 
Overall parameter set of the model is $\theta$. Consider one of the \textbf{fix target categories $t\in C$}, and the corresponding input $X^t$ (i.e., images that belong to the class $t$). $\hat{Y}_{C\setminus t}$ is the predicted output of non-target classes with distribution $q_{C\setminus t}=softmax(z_{C\setminus t})$, where $z_{C\setminus t}=[z_1, z_2, \dots, z_{t-1}, z_{t+1}, \dots, z_n]$. Note that all the random variables $y, z, q, t$ are dependent on input $x$, here we omit the dependence for brevity.

The problem formulation can be written as minimizing the MI between predicted output of non-target classes $\hat{Y}_{C\setminus t}$ and extracted features $\Phi(X^t)$, i.e. $I_{\theta}(\hat{Y}_{C\setminus t};\Phi(X^t))$. However, this MI cannot be computed in practice as the distribution of $\Phi(X^t)$ is intractable. To resolve the intractability issue, we apply the data processing inequality \cite{information_theory} and use the obtained upper bound, the MI between input $X^t$ of class $t$ and predicted output of non-target classes $\hat{Y}_{C\setminus t}$, as our objective:
\begin{equation}
    I_{\theta}(\hat{Y}_{C\setminus t};X^t),
    \label{equ:objective1}
\end{equation}
where $I_{\theta}(\hat{Y}_{C\setminus t};X^t)$ represents the mutual information under model parameter $\theta$.

According to the property of MI, i.e., $I(A;B)= H(A) - H(A|B)$, where $H$ is the entropy, and $A, B$ are two random variables, Eq. \ref{equ:objective1} can be decomposed into the difference between entropy and conditional entropy:
\begin{align}\nonumber
   I_{\theta}(\hat{Y}_{C\setminus t};X^t) = H_{\theta}(\hat{Y}_{C\setminus t}) - H_{\theta}(\hat{Y}_{C\setminus t}|X^t),
\end{align}
where $H_{\theta}(\cdot)$ denotes the entropy under model parameter $\theta$. Computing $H_{\theta}(\hat{Y}_{C\setminus t})$ involves the marginalization over $X^t$ which is computationally intractable in practice. Thus, instead of directly optimizing Eq. \ref{equ:objective1}, we consider optimizing its upper bound. From the fact that entropy reaches its upper bound when all the probabilities are equal, we have:
\begin{align*}\nonumber
    H_{\theta}(\hat{Y}_{C\setminus t}) \leq -(n-1)\frac{1}{n-1}\log\frac{1}{n-1}  = \log{(n-1)},
\end{align*}

The following lemma gives a theoretic justification for using the upper bound to replace Eq. \ref{equ:objective1}.
\begin{lemma}\label{lemma1}
    When the conditional probability distribution over non-target classes is uniform, the MI in Eq. \ref{equ:objective1} is 0, and hence $I_{\theta}(\hat{Y}_{C\setminus t};\Phi(X^t))=0$.
\end{lemma}

Lemma \ref{lemma1} shows that making the conditional distribution of the non-target classes uniform is desired. Thus we formulate the problem using the upper bound as maximizing $H_{\theta}(\hat{Y}_{C\setminus t}|X^t)$, which promotes the distribution to be uniform. When the mutual information is $I_{\theta}(\hat{Y}_{C\setminus t};\Phi(X^t))=0$, this suggests that the extracted features contains no information about the non-target classes.

In practice, we maximize the empirical conditional entropy for each class $t$:
\begin{align}\nonumber
    \frac{1}{N^t} \sum_{i=1}^{N_t} H_{\theta}(\hat{Y}_{C\setminus t}|x^t_i),
\end{align}
where $N_t$ is the number of training samples in class $t$.

We use the empirical conditional entropy as a regularization term with a weight parameter $\lambda$ into the CE objective function to form the overall objective function (i.e., \textbf{MaxNTE}):
\begin{align}
    \mathcal{L}_{MaxNTE} = {E}_{x\sim\mathcal{X}}[ D_{CE}(y||q(x;\theta)) - \lambda H_{\theta}(\hat{Y}_{C\setminus t}|X^t)].
    \label{equ:objective3}
\end{align}

Here, $H_{\theta}(\hat{Y}_{C\setminus t}|X^t)$ has a reachable upper bound. However, directly taking this as the objective function may not be the best choice for our goal, since the gradients become extremely small when closing to the upper bound. From Lemma \ref{lemma1}, we know a sufficient condition for $I_{\theta}(\hat{Y}_{C\setminus t};\Phi(X^t))=0$ is to enforce the conditional distribution of non-target classes to be uniform. In the following, we propose an efficient minimax loss based on game theory to achieve this target, which is lightweight and also insensitive to the choice of hyper-parameter. The comparison of MaxNTE and the new proposed loss will be provided in the experiments.
\begin{figure*}[t]
    \centering\centerline{\includegraphics[width=0.75\linewidth]{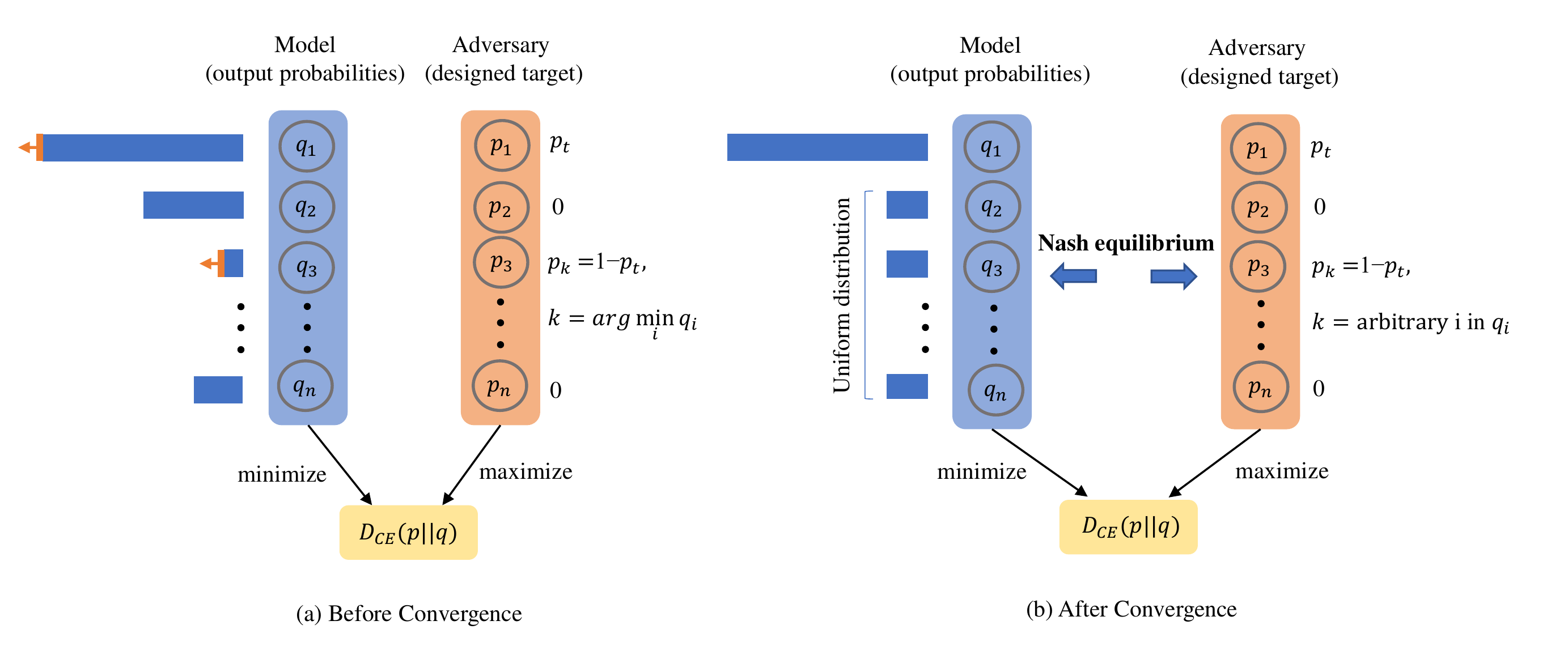}}
    \caption{(a) During the training phase (before convergence), the model keep promoting the smallest probability in the non-target model output by assign $1-p_{t}$ to the index of the minimum value in $q_{C\setminus t}$, i.e.., $q_{3}$ in (a). (b) After iterations of training, the model output distribution will finally be uniformly distributed over non-target classes. When reaching a convergence, a Nash equilibrium exists between the optimal solution of the model and the adversary.}
    \label{fig:overview}
\end{figure*}
\subsection{Game Theoretic Framework}
In this section, we introduce the game theoretic framework in detail. The resulting loss function will be shown in Eq. \ref{eq:MPP}, which is used as the major part of our main method in the experiments.
\subsubsection{Preliminaries}
Here we introduce some basic game-theoretic definitions \cite{myerson1991game} that we will use later.

\begin{definition} \label{game}
A strategic game is a tuple $G= \left \langle I,(A_{i})_{i\in I},(u_{i})_{i\in I} \right \rangle$, where $I$ is a nonempty set of players, $A_{i}$ is the set of actions available to each player $i\in I$, $A = \prod_{i \in I}A_i$ is the profiles of actions and $u_{i}: A \rightarrow R$ defines the payoff function for each player $i\in I$.  
 A two player strictly competitive game or zero-sum game is the strategic game $G$ with $I=\{1, 2\}$ and for all $a \in A$:
\begin{equation*}
    u_{1}(a) = -u_{2}(a).
\end{equation*}
\end{definition} 

\begin{definition}
   A mixed strategy set $S_i$ is the set of all probability distributions over $A_i$. $s_i \in S_i$ defines a mixed strategy for each player $i \in I$, and $s_i(a_i^k)$ is the probability that player $i$ plays $a_i^k \in A_i$. In a two player game, the expected payoff of player $i$ playing a mixed strategy against pure strategy $a_{-i}^*$ can be calculated as:
   \begin{equation*}
       U_i(s) = \sum_{a_i^k\in A_i} s_i(a_i^k) u_i(a_i^k, a_{-i}^*).
   \end{equation*}
   Likewise, the expected payoff of playing a pure strategy $a_i^*$ against mixed strategy can be calculated as:
   \begin{equation*}
       U_i(s) = \sum_{a_{-i}^k \in A_{-i}} s_{-i}(a_{-i}^k) u_i(a_i^*, a_{-i}^k),
   \end{equation*}
   where $-i$ denotes the player other than $i$.
\end{definition}

\begin{definition} \label{response}
Let $G$ be the strategic game, $i \in I$ be a player, and $s_{-i} \in S_{-i}$ be a strategy profile of players other than $i$. Then a strategy $s^{*}_i \in S_i$ is a \textit{best response} of player $i$ to $s_{-i}$ if:
\begin{equation*}
    \forall s_{i} \in S_{i}, \ \ U_{i}(s_{i}^{*}, s_{-i})\geq U_{i}(s_{i}, s_{-i}).
\end{equation*}
\end{definition}

\begin{definition} \label{nash}
A \textit{Nash equilibrium} of the strategic game $G$ is a action profile $s^{*} \in S$ such that for every player $i$, $s^{*}_{i}$ is the best response to $s^{*}_{-i}$.
\end{definition}

\subsubsection{$p-q$ zero-sum strategic game}
We define a zero-sum strategic game played between the \emph{model} and a designed \emph{adversary} with loss of the model defined as $D_{CE} (p||q)$. 
 We let $p$ be the ground truth label vector, normally one-hot encoding, while in our work we specifically design it for the objective.
Here, $p$ is the strategy of the adversary and $q$ is the strategy of the model. We assume that the model is a classifier with confidence $q_{t}$ on the target class. The model aims to assign the rest of the probability $1-q_{t}$ to non-target classes to minimize the loss. The adversary is the controller of the ground truth with fixed $p_{t}$ for the target class, and it aims to maximize the loss via adjusting the distribution on non-target class of $p$. This is a dynamic game that the two players play in order, in which the model goes first, and the adversary can adjust the strategy according to the previous action of the model. Fig. \ref{fig:overview} gives an overview of the proposed game theoretic framework.

\begin{definition}
For $p_t, q_t \in (0, 1)$, the defined strategic game is a tuple $G=\left \langle(P,Q),(A_P,A_Q), (u_P,u_Q)\right \rangle$ with:
\begin{align}\nonumber
    A_P &= \{(p_i)_{1\leq i \leq n}, p_i \in (0,1), \sum_{i \neq t}p_i = 1-p_t \}.\\\nonumber
     A_Q& = \{(q_i)_{1\leq i \leq n}, q_i \in (0,1), \sum_{i \neq t}q_i = 1-q_t \}.\\\nonumber
     u_P&=D_{CE} (p||q) = -u_Q.
\end{align}
\end{definition}

By Definition \ref{game}, $G$ is a two-player zero-sum game. In the following, we will prove the existence of Nash equilibrium.
\begin{theorem} \label{adversary}
For $p_t, q_t \in (0, 1)$, we have:
\begin{equation}\nonumber
  \forall q \in A_Q, \quad p^* = \mathop{\arg\max}_{p} D_{CE}(p||q),
\end{equation}
where $p^* = (p_i^*)_{1\leq i \leq n}$: 
\begin{equation}\nonumber
      p_i^* =
      \begin{cases}
        p_t, & i = t; \\
    	1-p_t, & i = k; \\
    	0, & otherwise
      \end{cases},
\end{equation}
for any $k =  \mathop{\arg\min}_z (q_{C\setminus t})_z$.

\end{theorem}

Note that in the rest of the paper, if there exists more than one minimum value in $q_{C\setminus t}$, $k =  \mathop{\arg\min}_z (q_{C\setminus t})_z$ refers to randomly taking one of them.

\begin{theorem} \label{model}
For $p_t, q_t \in (0, 1)$, we have:
\begin{equation}\nonumber
    \quad q^* = \mathop{\arg\min}_{q \in A_Q}D_{CE}(p^*|| q),
\end{equation}
where
\begin{equation}\nonumber
  q_i^* =
    \left\{\begin{matrix}
    q_t, & i = t; \\ 
    \frac{1-q_t}{n-1}, & otherwise.
    \end{matrix}\right.
\end{equation}
\end{theorem}

Theorem \ref{adversary} gives the worst case payoff for the model $q$. However, since $p^*$ depends on the index of the minimum value in $q$, they are not the best responses to each other. For example, when $q=q^*$, the adversary chooses one of the indexes of the minimum values in $q$ to determine $p^*$. Once $p^*$ is fixed, $q$ is no longer the best responses to $p^*$, since there exist a better $q$ to get a higher payoff (e.g., change the position of the minimum value). Thus, we need randomize $p$ to avoid this situation. Specifically, when $q=q^*$, the model uniformly distributes the probabilities over non-target classes. The adversary can randomly choose one of them since they are all the smallest value and the adversary's strategy becomes a mixed strategy. In this case, the two strategies form a Nash equilibrium. To show this mathematically, we give the following theorem.
\begin{theorem}
    Define an action subset for the adversary:
    \begin{align*}\nonumber
        a_P^* \subset A_P = \left\{ (p_i) |  p_i =
                                      \begin{cases}
                                        p_t, & i = t; \\
                                    	1-p_t, & i = k; \\
                                    	0, & otherwise
                                      \end{cases}
                                     \right\},
    \end{align*}
    where $k=\{1, 2, \dots,t-1, t+1, \dots, n$\}. 
    
    For the model:
    \begin{equation}\nonumber
        a_Q^* = \left\{(q_i) |  q_i =
                        \left\{\begin{matrix}
                                q_t, & i = t; \\ 
                                \frac{1-q_t}{n-1}, & otherwise
                        \end{matrix}\right.
                        \\
    \right\}.
    \end{equation}
    Then we have the following strategies: 
    \begin{equation}\nonumber
        s_P^*(p) =
                    \left\{\begin{matrix}
                            \frac{1}{n-1}, & p \in a_P^*; \\ 
                            0, & otherwise.
                    \end{matrix}\right.
    \end{equation}
    \begin{equation}\nonumber
        s_Q^*(q) =
                    \left\{\begin{matrix}
                            1, & q \in a_Q^*; \\ 
                            0, & otherwise,
                    \end{matrix}\right.
    \end{equation}
    such that $s^* = (s_P^*, s_Q^*)$ forms a Nash equilibrium.
    \label{the:theorem 3}
\end{theorem}
The detailed proofs of the three theorems are in Appendix.

\subsubsection{Minimax Loss}
The Nash equilibrium in a two player zero-sum game is equivalent to a minimax solution \cite{ferreira2012minimax}. Thus, by training with the worst-case payoff $D_{CE}(p^*||q)$, we expect that the model output ultimately converges to the best response $q^*$.
Finally our proposed \textbf{minimax loss (MM)} is defined as: 
\begin{align}\label{eq:MPP}
    \mathcal{L}_{MM} 
    &= \mathbb{E}_{x\sim\mathcal{X}}[ D_{CE}(p^*||q)]  \nonumber\\
    &= \mathbb{E}_{x\sim\mathcal{X}}[-p_t\log q(x;\theta)_t -(1-p_t)\log q(x;\theta)_k],
\end{align}

where $k =  {\mathop{\arg\min}_{a}} (q_{C\setminus t})_a$.
Here, we leave $p_t$ as a hyper-parameter to weight the regularizer of the objective corresponding to the class $t$. When $p_{t}$ is set to 1 for all classes, the loss function is equivalent to the standard CE loss. It is worth noting that, in regularization methods such as label smoothing and confidence penalty, the number of log operations increases with the number of categories. MM has only one more log operation than the cross entropy loss, while gains more performance in many tasks.

\begin{figure}[tb]
    \centering
    \includegraphics[width=\linewidth]{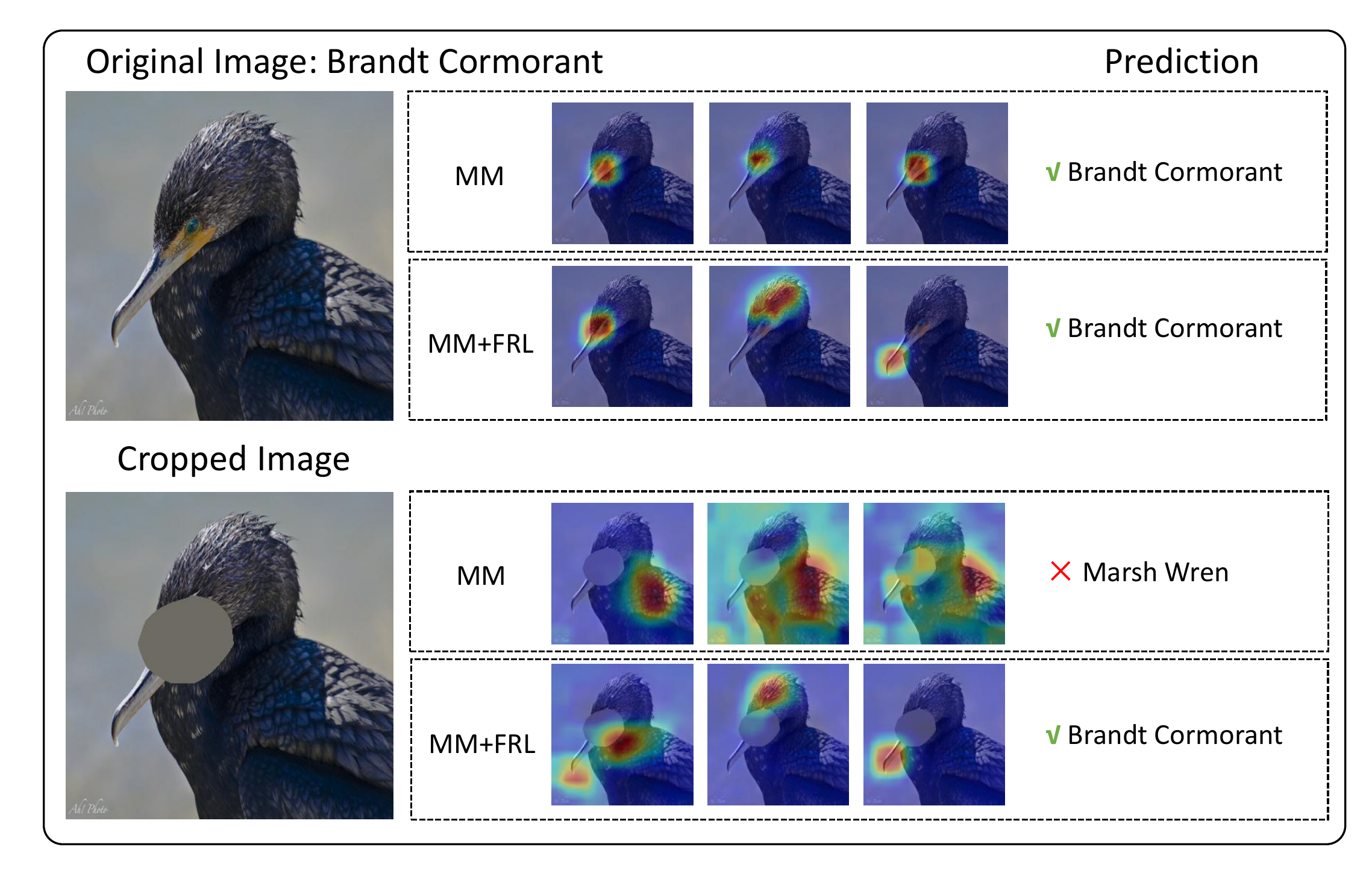}
    \caption{The feature redundancy problem in MM trained model. As shown in the first row, the three channels with the largest activation values detect the same region of the bird. That means the model's prediction is over dependent on the single feature. Once the region is cropped, as shown in the image below, the model cannot correctly predict the object category. The introduced FRL is shown to be able to eliminate this problem.}
    \label{fig:redundancy}
\end{figure}

\subsection{Feature Redundancy Loss}
MM promotes the feature uniqueness of each category. There may exist more than one solutions that having this property. In some cases, the obtained features can be redundant, different feature maps of a specific category are almost the same, as shown in the first row of Fig. \ref{fig:redundancy}. We want the extracted features to be more diverse, because single feature can be unreliable especially when the training set is small. Combining multiple features for decision can avoid wrong prediction under unexpected cases such as occlusion and make the model more robust. The second row of Fig. \ref{fig:redundancy} shows the case that one of the important regions is blocked, in which MM trained model that rely on single feature fails to make a correct prediction. Therefore we add an additional regularization term to choose more diverse features while maintaining the class uniqueness.

To enforce the difference of feature maps, we use normalized inner product to measure the similarity among feature maps of top activation as a loss function, named Feature Redundancy Loss. Specifically, in each forward sample, we select the feature maps that has top $K$ activation values before global average pooling. Let the shape of the selected feature maps $\phi = \Phi(x)$ from a sample $x$ be $(K, H, W)$, we calculate the normalized inner product between each pairs.
\begin{equation}
     \mathcal{L}_{FRL} = \sum_{i=1}^{K-1}\sum_{j=i+1}^{K} \frac{\langle\phi_i\phi_j\rangle}{\Vert{\phi_i}\Vert\Vert{\phi_j}\Vert}
\end{equation}
The loss can be calculated parallelly using tensor operation, thus a simple yet efficient trick. Our final loss is the weighted sum of MM and FRL:
\begin{equation}
    \mathcal{L} = \mathcal{L}_{MM} + \lambda \mathcal{L}_{FRL}
\end{equation}

\section{Experiments}
\subsection{Experimental Setup}
For evaluating our method, we use the following three benchmarks: CUB-200-2011 \cite{wah2011caltech}, FGVC-Aircraft \cite{maji2013fine}, Stanford Cars \cite{KrauseStarkDengFei-Fei_3DRR2013}. Further more, we assesses the effect of our method on standard visual classification benchmarks: CIFAR-10 \cite{krizhevsky2009learning}, CIFAR-100 \cite{krizhevsky2009learning}, STL-10 \cite{coates2011analysis}. Different methods are compared using ResNet18 \cite{he2016deep}, VGGNet11 \cite{simonyan2014very}, DenseNet161 \cite{huang2017densely} as the backbone models. The statistics of six datasets and the implementation details are introduced in Appendix.

\begin{figure}[t]
    \centering
    \subfigure[]{
        \centering
        \includegraphics[width=1.5in]{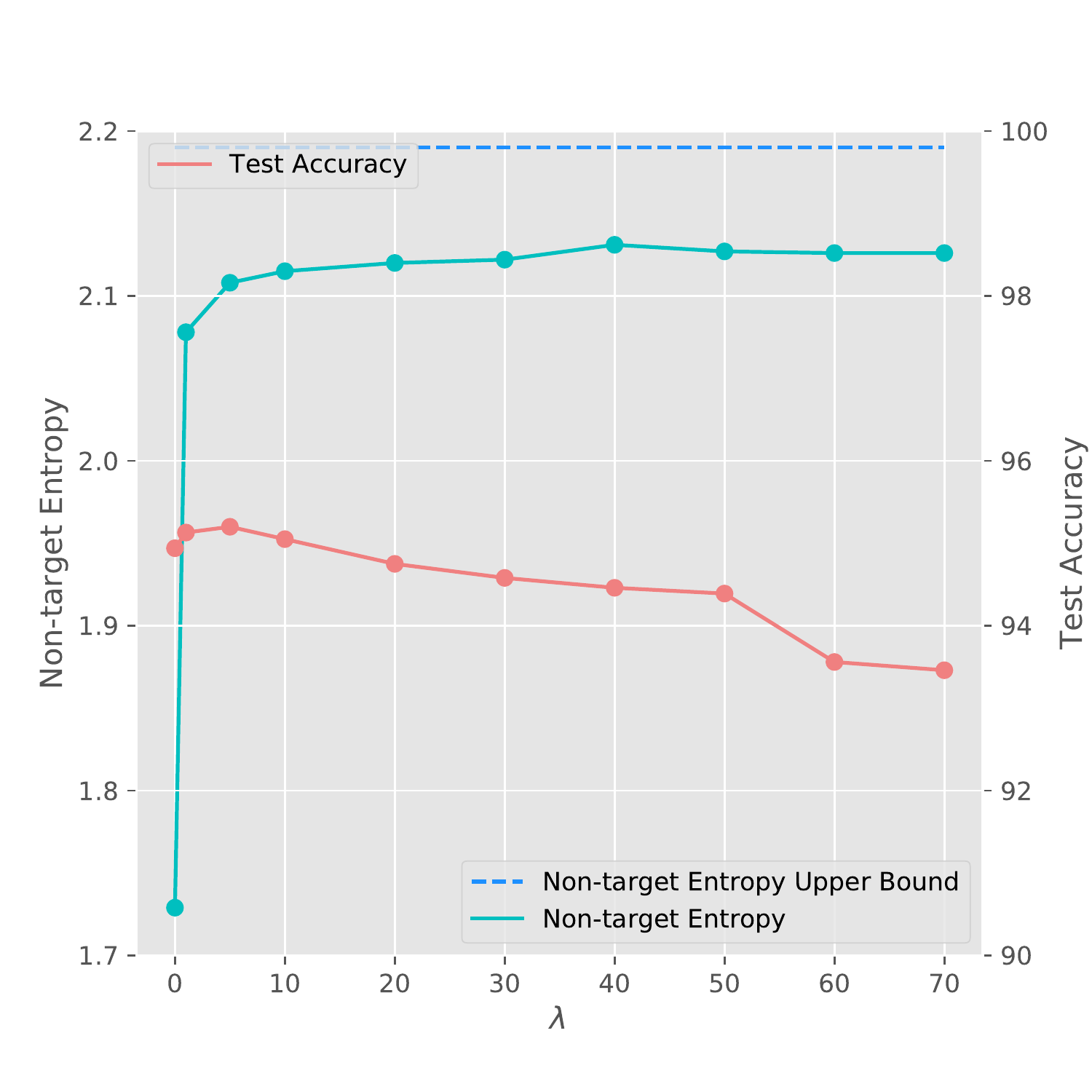}
    }
    \hspace{1mm}
    \subfigure[]{
        \centering
        \includegraphics[width=1.5in]{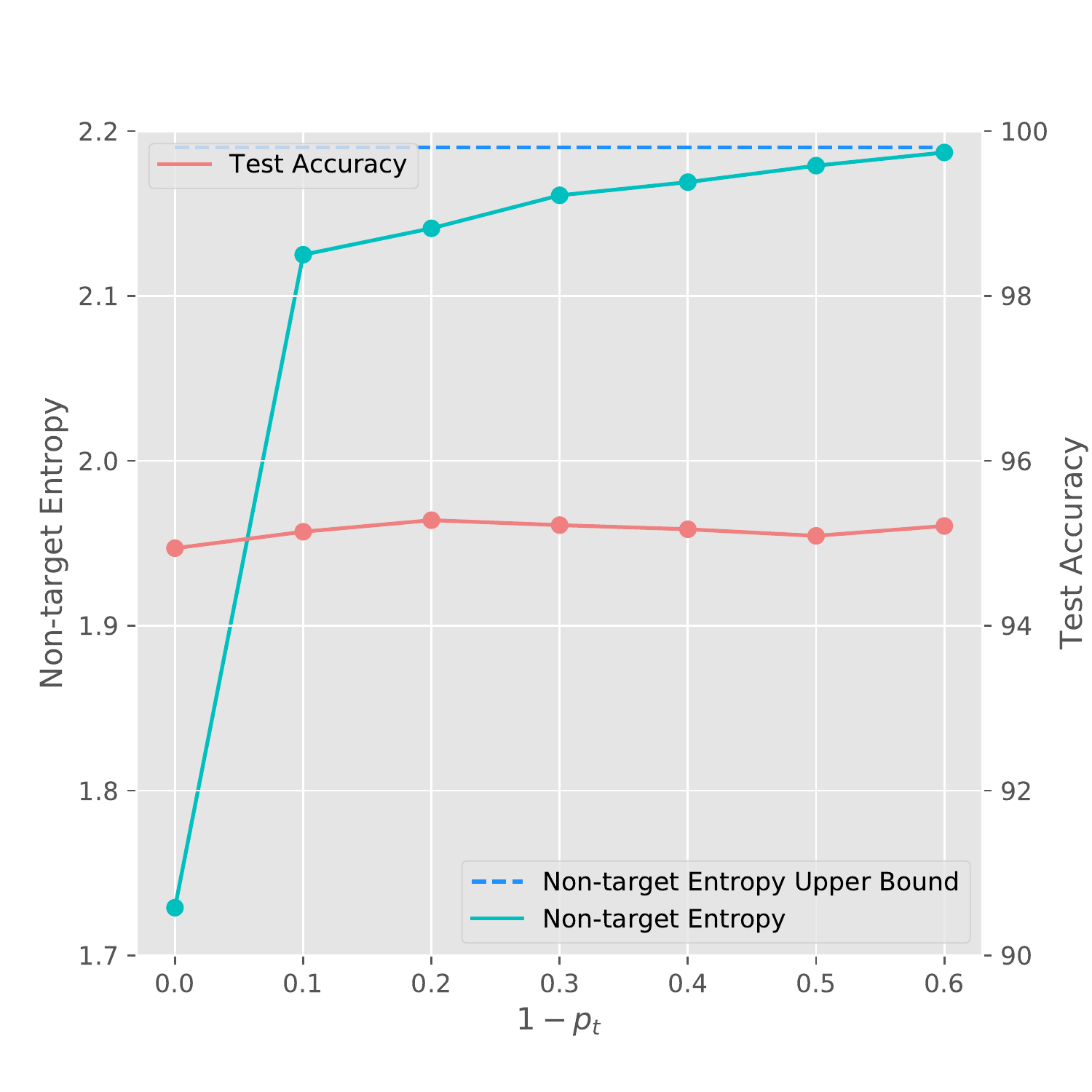}
    }
    \caption{Comparison between MM trained and MaxNTE loss trained models. Non-target classes entropy (scale using left vertical axis) and test accuracy (scale using right vertical axis) are shown. The entropy upper bound 2.197 is shown as the blue dotted line. We see that MM trained model reaches this upper bound, while MaxNTE trained model does not.}
    \label{fig:comparison with MaxNTE}
\end{figure}

We first compare our proposed MM with MaxNTE. Then we quantitatively compare our proposed method with different methods on FGVC tasks  as well as the standard visual classification tasks. Finally we conduct visualization to further show the effect of our method.

\subsection{Comparison between MaxNTE and MM}
We compare MM with the MaxNTE that directly minimize Eq. \ref{equ:objective3} on CIFAR-10 dataset. Note that when $p_t$ set to 1 for all classes (i.e. $1-p_t$ set to 0) and $\lambda$ set to 0, both of the losses are equal to the cross entropy loss. In MaxNTE, as shown in Fig. \ref{fig:comparison with MaxNTE}, the left figure, the test accuracy is decreasing when $\lambda$ is increasing, the maximum entropy reaches its bottleneck at about 2.12. The right figure shows that as $1-p_t$ (hyperparameter) increases, the entropy gradually approaches its upper bound ($\log(n-1)=\log9\approx2.19$) with stable test accuracy. The results show that MM can achieve our goal in a more efficient way, thus we apply it to the following experiments.

\subsection{Quantitative results}

\paragraph{Fine-Grained Visual Classification} From Table \ref{comparison with baseline model}, our proposed method improves the performance of three baseline models (i.e., ResNet-18, VGGNet-11 and DenseNet-161) across all three datasets (i.e., CUB-200-2011, FGVC-Aircraft and Stanford Cars). For example, training VGGNet-11 with MM obtained significant improvements of 2.50\% on average across three datasets compared with CE. LS performs better than CE since it also encourages the model to produce an output close to our objective. Besides, DenseNet-161 with MM achieves best results compared with other baselines.
\begin{table}[b]
\centering
\scriptsize
\caption{Comparison with three baseline models.}
\label{comparison with baseline model}
\begin{tabular}{c|c|c|c|c}
    \hline
    \hline
    Backbone & Method & CUB & Aircraft & Cars \cr
    \hline
    \hline
    \multirow{3}{40pt}{\centering {ResNet-18}} 
    & CE & 81.32 $\pm$ 0.31 & 89.89 $\pm$ 0.14 & 88.50 $\pm$ 0.21 \cr
    & LS & 81.83 $\pm$ 0.22 & 89.77 $\pm$ 0.27 & 91.06 $\pm$ 0.18 \cr
    & MM & \bf{83.14 $\pm$ 0.18} & \bf{90.37} $\pm$ 0.14 & \bf{91.74 $\pm$ 0.11} \cr
    \hline
    \multirow{3}{40pt}{\centering {VGGNet-11}} 
    & CE & 77.76 $\pm$ 0.28 & 85.38 s$\pm$ 0.66 & 87.32 $\pm$ 0.47 \cr
    & LS & 77.94 $\pm$ 0.23 & 87.57 $\pm$ 0.19 & 89.46 $\pm$ 0.32 \cr
    & MM & \bf{80.41 $\pm$ 0.15} & \bf{87.72}  $\pm$ 0.22 & \bf{89.83 $\pm$ 0.28} \cr
    \hline
    \multirow{3}{40pt}{\centering {DenseNet-161}} 
    & CE & 86.69 $\pm$ 0.32 & 90.94 $\pm$ 0.15 & 94.21  $\pm$ 0.12 \cr
    & LS & 87.63 $\pm$ 0.15 & 92.65 $\pm$ 0.21 & 94.27 $\pm$ 0.16 \cr
    & MM & \bf{87.98 $\pm$ 0.14} & \bf{93.34 $\pm$ 0.19} & \bf{94.72 $\pm$ 0.11} \cr
    \hline
    \hline
\end{tabular}
\end{table}

The overall experimental results compared with recent works including SOTA are shown in Table \ref{comparison with sota}. Our methods outperforms regularization-based methods (e.g. MaxEnt and PC) across all three datasets. While SOTA models achieve excellent results, they rely on extra structure or computational cost. Our proposed methods can fully bring out the potential of the baseline models and achieve SOTA in FGVC-Aircraft and Stanford Cars by only regularization. 

\begin{table}[t]
\centering
\scriptsize
\caption{Comparison with SOTA methods. * means the best performance among regularization-based methods.}
\label{comparison with sota}
\begin{tabular}{c|c|c|c}
    \hline
    \hline
    Method & CUB & Aircraft & Cars \cr
    \hline
    \hline
    B-CNN (\cite{lin2015bilinear})      & 84.1 & 84.1 & 91.3 \cr
    CBP (\cite{gao2016compact})         & 84.3 & 84.1 & 91.2 \cr
    KP (\cite{cui2017kernel})           & 86.2 & 86.9 & 92.4 \cr
    iSQRT-COV (\cite{li2018towards})    & 88.7 & 91.4 & 93.3 \cr
    MA-CNN (\cite{zheng2017learning})   & 86.5 & 91.8 & 92.8 \cr
    RA-CNN (\cite{fu2017look})          & 85.3 & 92.5 & 93.0 \cr
    MAMC (\cite{sun2018multi})          & 86.5 & ---  & 93.0 \cr
    DFL-CNN (\cite{wang2018learning})   & 87.4 & ---  & 93.8 \cr
    NTS-Net (\cite{yang2018learning})   & 87.5 & 91.4 & 93.9 \cr
    MaxEnt (\cite{dubey2018maximum})    & 86.5 & 89.2 & 92.9 \cr
    PC     (\cite{dubey2018pairwise})   & 86.9 & 89.8 & 93.0 \cr
    DCL (\cite{chen2019destruction})    & 87.8 & 93.0 & 94.5 \cr
    S3N (\cite{ding2019selective}       & 88.5 & 92.8 & 94.7 \cr
    DF-GMM (\cite{wang2019weakly}       & 88.8 & 93.8 & 94.8 \cr
    MGE-CNN (\cite{zhang2019learning}   & 89.4 & ---  & 93.9 \cr
    GCL (\cite{wang2020graph}           & 88.3 & 93.2 & 94.0 \cr
    API-Net (\cite{zhuang2020learning}) & \bf{90.0} & 93.9 & \bf{95.3} \cr
    ELoPE (\cite{hanselmann2020elope}   & 88.5 & 93.5 & 95.0 \cr
    DFL (\cite{liu2020filtration}       & 89.1 & 93.4 & 94.3 \cr
    CIN (\cite{gao2020channel})         & 88.1 & 92.8 & 94.5 \cr
    ACNet (\cite{ji2020attention}       & 88.1 & 92.4 & 94.6 \cr
    \hline
    DenseNet161+MM(Ours)                & 88.0 & 93.3 & 94.7 \cr
    DenseNet161+MM+FRL(Ours)            & \bf{88.5}* & \bf{94.0} & \bf{95.2}* \cr
  \hline
\hline
\end{tabular}
\end{table}

\begin{table}[b]
\centering
\scriptsize
\caption{Comparison with three baseline models on standard visual classification tasks.}
\label{comparison with standard vc}
\begin{tabular}{c|c|c|c|c}
    \hline
    \hline
    Backbone & Method & CIFAR-10 & CIFAR-100 & STL-10 \cr
    \hline
    \hline
    \multirow{4}{40pt}{\centering {VGGNet-11}} 
    & CE  & 92.26$\pm$0.08 & 70.37$\pm$0.33 & 79.80$\pm$0.31 \cr
    & CP  & \bf{92.62$\pm$0.05} & 70.30$\pm$0.19 & 80.17$\pm$0.14 \cr
    & LS  & 92.28$\pm$0.06 & 71.34$\pm$0.07 & 80.41$\pm$0.08 \cr
    & MM & 92.43$\pm$0.06 & \bf{71.62$\pm$0.18} & \bf{82.26$\pm$0.09} \cr
    \hline
    \multirow{4}{40pt}{\centering {ResNet-18}} 
    & CE  & 94.94$\pm$0.12 & 75.79$\pm$0.03 & 83.44$\pm$0.23 \cr
    & CP  & 95.11$\pm$0.01 & 76.01$\pm$0.31 & 83.75$\pm$0.02 \cr
    & LS  & 95.08$\pm$0.11 & 76.24$\pm$0.21 & 84.03$\pm$0.01 \cr
    & MM & \bf{95.33$\pm$0.12} & \bf{76.64$\pm$0.07} & \bf{85.42$\pm$0.04} \cr
    \hline
    \hline
\end{tabular}
\end{table}

\paragraph{Standard Visual Classification} We compare our method with two output regularization based methods: Confidence Penalty (CP)\cite{pereyra2017regularizing} and Label Smoothing (LS)\cite{szegedy2016rethinking}. As shown in Table \ref{comparison with standard vc}. our method outperforms several output regularization based methods across almost all the datasets and architectures. In CIFAR-10 and CIFAR-100, the improvements are not significant, and the test accuracy of VGGNet-11 with confidence penalty is slightly higher than that of our method in CIFAR-10. In STL-10, our method outperforms three baselines by a large margin.

\paragraph{Ablation Study}
We perform ablation experiments to show how different parts of our method work. As shown in Table \ref{ablation}, both MM and FRL greatly improves the performance of the baseline model. MM brings greater improvement overall, because CE can not precisely encode the unique feature even with FRL. Moreover, as FRL works by reducing the similarity of different feature maps, the effect of FRL on baseline model shows that feature redundancy also exists in regular training using CE. It is worth noting that hyper-parameters $\lambda$, $K$ and $p_t$ have little affect on the proposed MM and FRL.

\begin{table}[b]
\centering
\scriptsize
\caption{Ablation study using DenseNet-161 as the baseline model.}
\label{ablation}
\begin{tabular}{c|c|c|c|c}
    \hline
    \hline
    Minimax & FRL & CUB & Aircraft & Cars \cr
    \hline
    \hline
    &  & 86.69 $\pm$ 0.32 & 91.26 $\pm$ 0.15 & 94.21 $\pm$ 0.12 \cr
    $\checkmark$ &  & 87.98 $\pm$ 0.14 & 93.34 $\pm$ 0.19 & 94.74 $\pm$ 0.11\cr
    & $\checkmark$ & 87.14 $\pm$ 0.16 & 92.71 $\pm$ 0.48 & 94.71 $\pm$ 0.04 \cr
    $\checkmark$ & $\checkmark$ & \bf{88.48 $\pm$ 0.24} & \bf{93.96 $\pm$ 0.11} & \bf{95.18 $\pm$ 0.14}\cr
    \hline
    \hline
\end{tabular}
\end{table}

\subsection{Qualitative Result}

\begin{figure}[t]
    \centering
    \includegraphics[width=\linewidth]{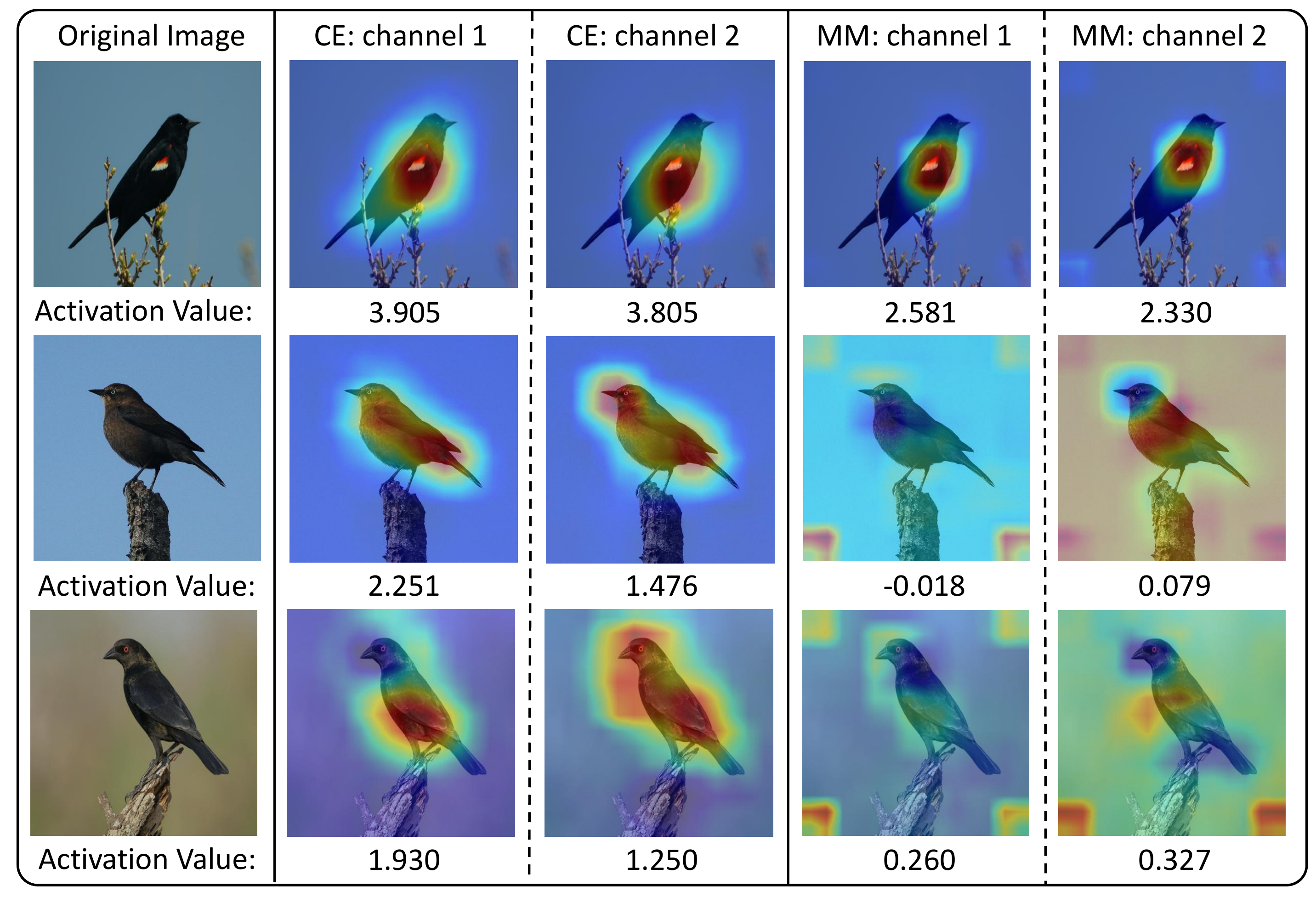}
    \caption{The comparison of the learned filters of models trained with CE and our MM. We select the three images of similar bird species to demonstrate the effect of our method. The two channels with the largest activation values according to the forward pass of Red Winged Blackbird (the images in the first row) are presented. Although CE trained filters are able to capture the important features, they are confused with other categories. MM trained filters can extract precisely the unique features which will not be activated when encounter similar objects.}
    \label{fig:cam_1}
\end{figure}

\begin{figure}[ht]
    \centering
    \includegraphics[width=\linewidth]{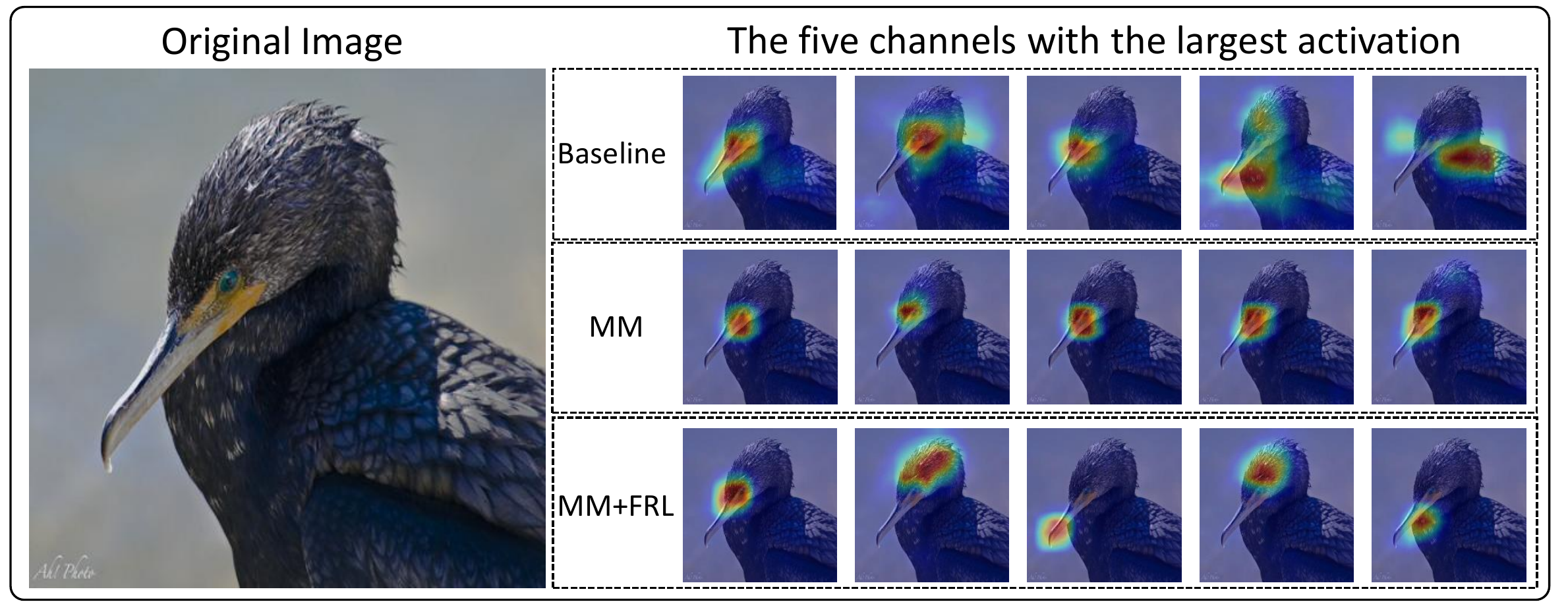}
    \caption{The comparison of the filters with largest activation values of different models. The left side present the original image. On the right side, the first row is the model trained with CE as the baseline. The second row is the model trained with MM and the third row is the model trained with MM and FRL. Compared with baseline, MM minimizes the activation of irrelevant regions, so as to produce a clear and focused activation map. On the basis of the effect of MM, FRL encourages the model to focus on different discriminative regions.}
    \label{fig:cam_2}
\end{figure}

To show in detail how our approach works, we visualize the penultimate layer feature maps with top activation values. We up-sample the feature maps to match the original image by bi-linear interpolation. As shown in Fig. \ref{fig:cam_1}, CE trained model (in the first row, column 2, 3) correctly localize the important parts of the bird, but it will cover irrelevant areas, which may lead to false triggering of other features (see the second and the third row). The filters trained by MM that responsible for detecting the unique features of Red Winged Blackbird do not response to other species as CE trained filters do. 

Fig. \ref{fig:cam_2} shows the second row demonstrate the different effects of our proposed MM and FRL. With our regularization, the model become more concentration so as to avoid introducing information about irrelevant categories. However, the regularization objective will lead to redundancy features, i.e. all the feature maps point on the same area, which means that the model become overdependent on single feature, and may be harmful for generalization. FRL well eliminates this problem by forcing the model to distract its attention. The final results are shown in the last row of Fig. \ref{fig:cam_2}, the model attention become both focused and diversified. With a clearer activation maps, we can better see how the model works.

\section{Conclusion}
In this work, we provide an information theoretic point of view, to address the major challenge in FGVC, i.e., learning the features unique to categories. We formulate the aim to minimizing the MI between the learned features and non-target classes, based on which we deduce an explicit regularization objective. To efficiently achieve our objective, we construct a game-theory based framework to derive a stable minimax loss, which is proved to converge to a Nash equilibrium. Furthermore, FRL is proposed to avoid over depending on single feature as a complement of MM. As a result, the model is able to extract the most distinctive parts of the object and reduce the influence of background noise. By only regularization, our proposed methods bring the potential of the baseline models into full play and achieves competitive results with SOTA models without extra computational cost.
\bibliography{arxiv}
\bibliographystyle{icml2021}

\end{document}


\twocolumn[
\icmltitle{Learning Class Unique Features in Fine-Grained Visual Classification: Appendix}



\icmlsetsymbol{equal}{*}

\begin{icmlauthorlist}
\icmlauthor{Aeiau Zzzz}{equal,to}
\icmlauthor{Bauiu C.~Yyyy}{equal,to,goo}
\icmlauthor{Cieua Vvvvv}{goo}
\icmlauthor{Iaesut Saoeu}{ed}
\icmlauthor{Fiuea Rrrr}{to}
\icmlauthor{Tateu H.~Yasehe}{ed,to,goo}
\icmlauthor{Aaoeu Iasoh}{goo}
\icmlauthor{Buiui Eueu}{ed}
\icmlauthor{Aeuia Zzzz}{ed}
\icmlauthor{Bieea C.~Yyyy}{to,goo}
\icmlauthor{Teoau Xxxx}{ed}
\icmlauthor{Eee Pppp}{ed}
\end{icmlauthorlist}

\icmlaffiliation{to}{Department of Computation, University of Torontoland, Torontoland, Canada}
\icmlaffiliation{goo}{Googol ShallowMind, New London, Michigan, USA}
\icmlaffiliation{ed}{School of Computation, University of Edenborrow, Edenborrow, United Kingdom}

\icmlcorrespondingauthor{Cieua Vvvvv}{c.vvvvv@googol.com}
\icmlcorrespondingauthor{Eee Pppp}{ep@eden.co.uk}

\icmlkeywords{Machine Learning, ICML}

\vskip 0.3in
]




\section{Appendix 1: Experimental Details}
Our experiments are performed on fine-grained visual classification (FGVC) benchmarks: CUB-200-2011 \cite{wah2011caltech}, FGVC-Aircraft \cite{maji2013fine}, Stanford Cars \cite{KrauseStarkDengFei-Fei_3DRR2013} and standard visual classification benchmarks: CIFAR-10 \cite{krizhevsky2009learning}, CIFAR-100 \cite{krizhevsky2009learning}, STL-10 \cite{coates2011analysis}. The statistics of six datasets are shown in \ref{statistics}.

Different methods are compared on ResNet18 \cite{he2016deep}, VGGNet11 \cite{simonyan2014very}, DenseNet161 \cite{huang2017densely}. We conduct all our experiments on Pytorch framework \cite{paszke2019pytorch} with NVIDIA 2080Ti GPUs. In all the datasets, we perform three-folds cross validation to choose the hyper-parameter (including the algorithm-specific parameters, learning rate, decay rate and weight decay, see Table \ref{training details}. Then we evaluate all the models over three runs and report the mean and standard deviation on the test set. We use Stochastic Gradient Descent (SGD) to update the model parameters. All the images are normalized and augmented by random crop and random horizontal flip. For algorithm-specific hyper-parameters, we set 1.00 for the weight of Confidence Penalty (CP), 0.10 for the smoothing rate of Label Smoothing (LS) and 0.85 for $p_t$ of Minimax Loss (MM). For FRL, we set $K=10$ and $\lambda=1$

\begin{table}[htbp]
\centering
\scriptsize
\caption{Statistics of six datasets in this work.}
\label{statistics}
\begin{tabular}{c|c|c|c}
    \hline
    Dataset & \#Training & \#Testing & \#Categories \cr
    \hline
    \hline
    CUB-200-2011  & 5994 & 5794 & 200 \cr
    FGVC-Aircraft  & 6667 & 3333 & 100 \cr
    Stanford Cars & 8144 & 8041 & 196 \cr
    \hline
    \hline
    CIFAR-10  & 50000 & 10000 & 10 \cr
    CIFAR-100  & 50000 & 10000 & 100 \cr
    STL-10 & 5000 & 8000 & 10 \cr
    \hline
    \hline
\end{tabular}
\end{table}

\begin{table*}[htbp]
\centering
\scriptsize
\caption{Training hyper-parameters for experiments. In LR policy, n1/n2: decay n1 for every n2 epochs.}
\label{training details}
\begin{tabular}{c|c|c|c|c|c|c|c}
    \hline
    \hline
    
    Dataset & Image sizet & Crop size & Batch size & Epochs & Learning rate & Weight decay & LR policy \cr
    \hline
   
    CIFAR-10 & 32$\times$32 & 32$\times$32 & 128 & 200 & 0.1 & 0.0005 & 0.2/50 \cr
    
    CIFAR-100 & 32$\times$32 & 32$\times$32 & 128 & 200 & 0.1 & 0.0005 & 0.2/50\cr
    
    STL-10 & 96$\times$96 & 96$\times$96 & 64 & 200 & 0.1 & 0.0005 & 0.2/50\cr
  
    \hline
    CUB-200-2011 & 512$\times$512 & 448$\times$448 & 16 & 60 & 0.004 & 0.0005 & 0.9/2\cr
    
    FGVC-Aircraft & 512$\times$512 & 448$\times$448 & 16 & 60 & 0.008 & 0.0005 & Cosine\cr
    
    Stanford Cars & 512$\times$512 & 448$\times$448 & 16 & 60 & 0.01 & 0.0005 & Cosine\cr
    
    \hline
    \hline
\end{tabular}
\end{table*}

    
   
    
    
    
    
    
  
    
    
    
    
    
    
    
    
    

\section{Appendix 2: The Proofs}

\subsection{Proof of lemma 1}

\begin{proof}
From previous discussions, we have
\begin{align}\nonumber
     I_{\theta}(\hat{Y}_{C\setminus t};X^t) \leq \log{(n-1)} - H_{\theta}(\hat{Y}_{C\setminus t}|X^t).
\end{align}
On the other hand, when the conditional probability distribution over non-target classes is uniform, we have 
\begin{equation}\nonumber
    H_{\theta}(\hat{Y}_{C\setminus t}|X^t)= \log(n-1). \nonumber
\end{equation}
Therefore $I_{\theta}(\hat{Y}_{C\setminus t};X^t)\leq 0$, combining with the fact that MI is always non-negative gives $I_{\theta}(\hat{Y}_{C\setminus t};X^t)=0$. Finally we have $I_{\theta}(\hat{Y}_{C\setminus t};\Phi(X^t))\leq I_{\theta}(\hat{Y}_{C\setminus t};X^t)=0$, and thus $I_{\theta}(\hat{Y}_{C\setminus t};\Phi(X^t))=0$.
\end{proof}

\subsection{Proof of theorem 1}
\begin{proof}
Since $k =  \mathop{\arg\min}_z (q_{C\setminus t})_z$, then $\forall i \neq k, q_i \geq q_k$. We have:
\begin{align*}\nonumber
    D_{CE}(p || q) &= 
      - \sum_{i\neq k,t} p_i \log q_i - p_k\log {q_k} - p_t\log q_t \nonumber\\
    &\leq - \sum_{i\neq k,t} p_i \log q_k - p_k\log {q_k} - p_t\log q_t \nonumber\\
    &=  - (\sum_{i \neq k,t}p_i + p_k)\log q_k - p_t\log q_t \nonumber\\
    &=  - (1-p_t) \log q_k - p_t\log q_t \nonumber\\
    &=  D_{CE}(p^*|| q), \nonumber
\end{align*}
which concludes Theorem 1.
\end{proof}

\subsection{Proof of theorem 2}
\begin{proof}
Let $k =  \mathop{\arg\min}_z (q_{C\setminus t})_z$. Since $\forall i \neq t, q_i^* = \frac{1-q_t}{n-1}$, for any $q \neq q^*$, we have $\forall i \neq t, q_i^* \geq q_{k}:$
\begin{align*}\nonumber
    D_{CE}(p^*|| q) 
    &=  -\sum_i p_i^*\log q_i \nonumber\\
    &=  - (1-p_t)\log q_{k} - p_t\log q_t, \nonumber\\
  & \geq  - (1-p_t)\log q_i^* - p_t\log q_t, \nonumber 
  \label{equality}\\
  & =  D_{CE}(p^*||q^*),
\end{align*}
which concludes Theorem 2.
\end{proof}

\subsection{Proof of theorem 3}
\begin{proof}
    Under the strategy of $s^*$, the expected payoff of the adversary can be calculated as:
    \begin{align}
        U_P = \sum_{a_P \in a_P^*} s_P^*(a_P) u_P(a_P, a_Q), a_Q \in a_Q^*. \nonumber
    \end{align}
  Since ${\rm argmin}_z{(q_{C\setminus t})_z}$ contains all the indexes except $y$ under the condition that $Q$ uniformly distribute all the probabilities over non-target classes, thus $\forall p \in a_P, p = p*$. Moreover, $a_Q$ contains only one action which is equal to $q^*$, thus:
    \begin{align*}
            U_P &= n-1 \times \frac{1}{n-1} u_P(p^*, q^*) \nonumber
            = u_P(p^*, q^*). \nonumber
    \end{align*}
    Assume that there is another strategy $s_P$ that gives higher payoff for the adversary, then:
    \begin{align*}
        U_P' &= \sum_{a_P \in a_P} s_P^*(a_P) u_P(a_P, a_Q), a_Q \in a_Q^* \nonumber\\
                      &\leq \sum_{a_P \in a_P} s_P^*(a_P) u_P(p^*, a_Q), a_Q \in a_Q^*, \forall p^* \in a_P^* \nonumber\\
                      &= u_P(p^*, q^*) = U_P,
    \end{align*}
    which is contrary to the assumption, thus prove that the $s_P^*$ is the best response to $s_Q^*$.
    
    Likewise, we can prove that $s_Q^*$ is the best response to $s_P^*$. Thus we have proved that $s^* = (s_P^*, s_Q^*)$ forms a Nash equilibrium.
\end{proof}

\bibliography{example_paper}
\bibliographystyle{icml2021}